\documentclass[11pt]{article}

\usepackage{acl}
\usepackage[utf8]{inputenc}
\usepackage{CJKutf8}
\usepackage{xcolor}
\usepackage{arabtex}
\usepackage{colortbl}

\usepackage{times}
\usepackage{latexsym}

\usepackage{float}

\usepackage{booktabs}
\usepackage{float,lscape}
\usepackage{array}

\usepackage{makecell}

\usepackage{pgfplots}
\pgfplotsset{width=10cm,compat=1.9}

\usepackage{multirow}

\usepgfplotslibrary{external}
\usepackage[T1]{fontenc}
\usepackage{microtype}

\usepackage[arabic,USenglish]{babel}
\usepackage{inconsolata}
\usepackage{abstract}

\definecolor{lightgreen}{rgb}{0.67, 0.94, 0.82}
\definecolor{darkorchid}{rgb}{0.6, 0.2, 0.8}

\title{Organic Data-Driven Approach for Turkish Grammatical Error Correction and LLMs}

\author{As{\i}m Ersoy \hspace{1em} Olcay Taner Y{\i}ld{\i}z  \hspace{.5em}\\
  Özyeğin University \\ 
  \texttt{asem.okby@ozu.edu.tr}\hspace{.5em} \texttt{olcay.yildiz@ozyegin.edu.tr}\\ 
  }

\begin{document}
\maketitle

\begin{abstract}

Grammatical Error Correction has seen significant progress with the recent advancements in deep learning. As those methods require huge amounts of data, synthetic datasets are being built to fill this gap. Unfortunately, synthetic datasets are not organic enough in some cases and even require clean data to start with. Furthermore, most of the work that has been done is focused mostly on English. In this work, we introduce a new organic data-driven approach, clean insertions, to build parallel Turkish Grammatical Error Correction datasets from any organic data, and to clean the data used for training Large Language Models. We achieve state-of-the-art results on two Turkish Grammatical Error Correction test sets out of the three publicly available ones. We also show the effectiveness of our method on the training losses of training language models.
 
\end{abstract}

\section{Introduction}\label{sec1}

Humans naturally tend to make typos for various factors. Those typos and grammatical errors propagate to the data used in Natural Language Processing (NLP) systems and any data-related tasks, which could lead to unexpected behavior. For instance, a sentiment analysis text classifier that has been trained with a frequently occurring misspelled word may produce unexpected results when processing correctly spelled words in the input. Another example that we looked into closely is Large Language Models (LLMs) which are trained on massive amounts of data mostly from the internet such as the OSCAR dataset\footnote{https://huggingface.co/datasets/oscar}. We observe a significant percentage of grammatical mistakes in the OSCAR dataset, specifically, in the Turkish OSCAR data, which has an effect on the training losses and causes the models sometimes to generate erroneous text. These examples show the importance of the NLP task Grammatical Error Correction (GEC) in facilitating text-based communications.

Given the GEC task's importance, many works addressed the task and, with the advancement and rise of deep learning techniques, achieved significant progress on the task \citep{bryant2023grammatical}. Unfortunately, most of that work focused on English and some other common languages. On the other hand, the work done for Turkish is few and limited, which explains how the Turkish GEC task is barely noticed and paid attention to. There are only two open-source evaluation sets available with more than one error type \citep{koksal-etal-2020-turki, kara2023gecturk}. And, there is only one open-source synthetic training set with a pre-defined set of error types from \citep{kara2023gecturk} utilizing inorganic data such as newspaper data to build GEC datasets, which leads to poor performance on common general errors that an average typer could make. 

In this work, we tackle the Turkish GEC task with a new organic data-driven approach, addressing the problem of inorganic and artificial datasets utilized for GEC. We introduce a simple method that we call clean insertions. It involves building an incorrect-correct spelling dictionary to be used in replacing commonly made misspellings of words and phrases with their correct versions in any organic text, e.g. text crawled from the internet, which mostly contains grammatical errors. The spelling dictionary and its size are crucial and have a major effect on the produced dataset's quality. 

This method leads to a partially correct parallel text since the spelling dictionary probably does not contain all the existing mistakes in the dataset. Despite this fact, our simple method achieves state-of-the-art results on two different test sets out of the three available open-source evaluation sets. In addition to that, we use GPT-4 to automatically build a parallel GEC dataset and compare the models we train on those different datasets. Furthermore, we run experiments to test and show the effectiveness of our method on cleaning training data for LLMs. We open-source different datasets and models with this work for the Turkish GEC task. Here are our work's contributions:

\begin{itemize}

\item 
We introduce a new organic data-driven approach, clean insertions, to build synthetic GEC datasets from any organic data, which mostly contains grammatical mistakes. No clean data is required!

\item 
We find that partially corrected GEC datasets could be utilized to achieve state-of-the-art results. 

\item 
We find that cleaning the data used for training LLMs leads to lower loss values.

\item
We open-source\footnote{https://github.com/asimokby/Turkish-GEC} 1) A manually annotated spelling dictionary consisting of about 150k incorrect-correct word and phrase pairs. 2) The largest Turkish GEC parallel dataset consisting of 2.3m sentences. 3) A Turkish GEC dataset of about 100k sentences annotated by GPT. 4) The largest test set for Turkish GEC which consists of about 2,400 manually corrected sentences. 5) All the best-performing models trained in this work.

\end{itemize}

We structure our paper as follows: In Section \ref{sec2}, we follow up the introduction with a literature review of the work done on Grammatical Error Correction covering the datasets, approaches, and Turkish GEC. Then, in Section \ref{sec3}, we detail in the data methodology section the development of the OSCAR GEC and GPT GEC datasets and our clean insertions method. Later on, in Section \ref{sec4}, we touch on the experimental setup, the training, and the evaluation of our models trained on our datasets and other open-source datasets. In Section \ref{sec5}, we show the evaluation results for both correction and detection. In Section \ref{sec6}, we briefly touch on the language models and the effect of our method on the training losses of language models. Finally, we sum up the work with a conclusion in Section \ref{sec13}.

\section{Related Work}\label{sec2}

Several datasets and models have been developed to address the grammatical error correction task. We review some of those in this section: 

\subsection{Datasets}
\
Datasets that have been utilized for Grammatical Error Correction mostly consist of English academic essays authored by either English learners and native speakers \citep{yannakoudakis2012modeling, dahlmeier2013building, napoles2017jfleg, bryant2019bea}. Other datasets included web data such as in \citep{flachs2020grammatical} which contains random paragraphs sampled from the Common-Crawl dataset\footnote{https://commoncrawl.org/}. Some studies put some effort into filling the gap and built non-English datasets for less popular languages in NLP such as Arabic \citep{mohit2014first}, Chinese \citep{lee2018building}, and Turkish \citep{koksal-etal-2020-turki}.

In addition to the human-labeled datasets mentioned above, the advancements in deep learning surged the need for large synthetic datasets. Mainly, there are two techniques used in building such datasets: noisy injections and back-translation \citep{kiyono2019empirical}. The Noisy injections technique involves corrupting an already clean text by inserting some pre-defined errors in a rule-based way \citep{ehsan2013grammatical, lichtarge2019corpora, zhao2019improving}, or by injecting probabilistic error patterns \citep{rozovskaya2010training, felice2014generating, rei2017artificial}. The back-translation technique, on the other hand, involves training a noisy channel model to predict a probable source text given a correct text \citep{xie2018noising}.  

\subsection{Approaches}

Approaches used in Grammatical Error Correction developed over time, beginning with rule-based approaches \citep{naber2003rule} due to their straightforwardness. Later on, data-driven approaches emerged such as single-error-type classifiers \citep{lee2004automatic, chodorow2007detection, berend2013lfg, lee2008correcting}, where each classifier targets a specific error type independently, assuming the surrounding context is correct \citep{bryant2023grammatical}, which is a limitation of rule-based approaches. Statistical Machine Translation (SMT) approaches \citep{brockett2006correcting, mizumoto2011mining, yuan2013constrained} come into the picture to address the limitation of the rule-based approaches by correcting all error types simultaneously \citep{bryant2023grammatical}. SMT systems leverage statistical models trained on parallel corpora to generate translations by estimating the likelihood of different translations and selecting the most probable one, treating the GEC task as a translation task \citep{bryant2023grammatical}. The complexity of the SMT approaches, e.g. relying on separate translation and language models, is addressed by neural machine translation (NMT), which consists of a single neural network. 

NMT approaches, which achieve state-of-the-art results, are encoder-decoder methods \citep{cho2014learning} where encoders and decoders could be of different possible architectures such as RNNS \citep{bahdanau2014neural}, CNNS \citep{gehring2016convolutional}, or Transformers \citep{vaswani2017attention}, which were applied successfully on the GEC task \citep{yuan2016grammatical, yuan2019neural, junczys2018approaching}. Recent approaches, utilize pre-trained large language models and achieve state-of-the-art results \citep{rothe2021simple, tarnavskyi2022ensembling} by only fine-tuning them, solving the data bottleneck requirement for large networks.

\subsection{Turkish Grammatical Error Correction}

Turkish Grammatical Error Correction hasn't been paid as much attention as English GEC. For example, recent work, \citep{arikan2019detecting} builds a synthetic dataset considering only a single error type. Later, \citep{koksal-etal-2020-turki} proposed the first public benchmark dataset of manually annotated 2000 Turkish tweets covering different error types. Then, recently, \citep{kara2023gecturk} built and open-sourced the first Turkish GEC synthetic large dataset, by making noisy injections into clean newspaper data, covering 25 error types. Additionally, they released a manually curated test set of 300 movie reviews to the public.

\section{Data Methodology}\label{sec3}

\begin{figure*}[h]
\centering
\includegraphics[width=1.0\textwidth]{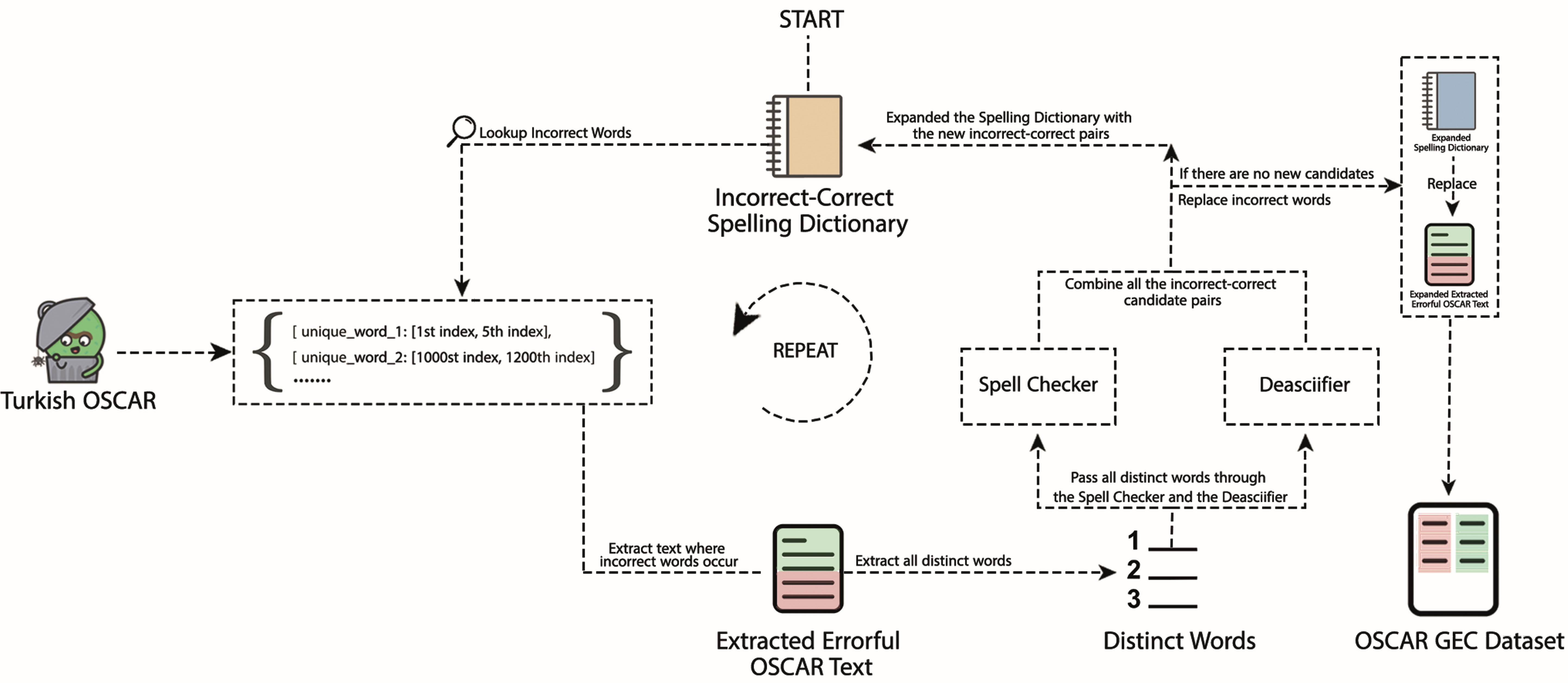}
\caption{A pipeline of the creation process of OSCAR GEC showing all the steps involved in creating the OSCAR GEC dataset}\label{fig:oscar_gec_pipeline}
\end{figure*}

We detail in this section the procedure followed in developing our datasets and give an overview of all the datasets utilized in this work. 

\subsection{OSCAR GEC}

Our data pipeline starts with building a manually annotated spelling dictionary of incorrect-correct 148,932 word and phrase pairs. To build this dictionary, we collected Turkish text from various sources and asked native Turkish speakers to extract incorrect words and write down their correct versions. We open-source with this work the spelling dictionary and its expanded version comprised of 703,938 pairs, which we introduce in the following sections. 

Given the manually created spelling dictionary, we expand our dictionary and build our OSCAR GEC dataset following the pipeline shown in Figure \ref{fig:oscar_gec_pipeline}. We first use the Turkish OSCAR Dataset to create a word-index dictionary, having each word in the corpus as a key and a list of indexes where the corresponding incorrect word occurs. Then, we apply the following steps: 1) Look up each incorrect word in our spelling dictionary in the word-index OSCAR dictionary. 2) Create a dataset of those texts we extract from OSCAR, each may contain one sentence or more. 3) Create a list of the unique words in the extracted OSCAR texts. 4) Run a word-level Deasciifier \ref{deasscii-sec} and Spell Checker \ref{checker-sec} to correct each word of the distinct words if possible. 5) Expand our spelling dictionary with these incorrect-correct pairs. The above steps are repeated until the increase in the size of the spelling dictionary stops.

Table \ref{iteration_details} shows the details of each iteration. In the first iteration, the manually created spelling dictionary is used. Starting from the 2nd iteration we notice an increase in the size of the spelling dictionary which leads to an increase in the size of the extracted OSCAR text. In the final iterations, the amount of increase in the spelling dictionary size starts to decrease until it becomes zero and the size of the spelling dictionary stabilizes and does not expand. Finally, all the text extracted from OSCAR is tokenized with a sentence tokenizer, merged, and deduplicated forming 2,326,921 sentences (or a text containing more than one sentence that couldn't be tokenized properly). Since we applied the tokenization later and some of the texts we extracted from OSCAR may composed of more than one sentence, some of those sentences do not necessarily contain any incorrect word or phrase and may be completely correct. 

To build our OSCAR GEC dataset, we apply our novel approach clean-insertions, which involves substituting incorrectly spelled words or phrases with their correct counterparts. We iterate over the final 2,326,921 sentences and for every word in each sentence, we perform a lookup in our expanded spelling dictionary of 703,938 incorrect-correct word pairs and replace the incorrect words with their counterparts. We end up with our OSCAR GEC dataset which contains 2,326,921 parallel sentences of incorrect-correct sentence pairs.

\subsubsection{Deasciifier}\label{deasscii-sec}

Deasciification is a process used in Turkish natural language processing (NLP) to convert text written in the Turkish language using ASCII characters into its proper form with Turkish-specific characters.

Turkish has specific characters such as "ı," "ş," "ğ," "ç," and "ü" that are not present in the standard ASCII character set. However, due to historical reasons, limitations of older computer systems, or simply out of habit, many texts written in Turkish may use ASCII characters as substitutes for these specific Turkish characters. For instance, "i" might be used instead of "ı," "s" instead of "ş," and so on.

Deasciification algorithms aim to detect and correct these substitutions, transforming the text into its correctly spelled Turkish form. We utilize a Deasciification algorithm\footnote{https://github.com/StarlangSoftware/TurkishDeasciifier} that works in the following steps: 1) Generates candidates of all the possible combinations of those characters in a word. 2) Uses a morphological analyzer to analyze each candidate version of the word. 3) Returns a candidate from those that pass the morphological analyzer i.e. are analyzable. However, we only include the words that have a single candidate to make sure that the candidate is indeed the correct version of the word.  

\subsubsection{Spell Checker}\label{checker-sec}

We make use of a Spell Checker\footnote{https://github.com/StarlangSoftware/TurkishSpellChecker} which generates a list of candidate words by performing various operations such as swapping adjacent letters, deleting letters, replacing letters with different characters, and adding new characters. The algorithm passes those candidates through a morphological analyzer and returns only the analyzable candidates, similar to the Deasciifier algorithm. And, again similar to the Deasciifier, we only consider the words that have one candidate. 

\begin{table*}[h]
\centering
\begin{tabular}{p{1.5cm}rrrr}
\toprule
Iteration & Spelling Dictionary & Extracted & OSCAR Distinct & Spelling Dictionary \\
Number & Size & OSCAR Texts & Words & Size Difference (+) \\
\midrule
1st    & 148,932   & 864,013  & 1,852,426 & - \\
2nd    & 463,072   & 1,220,251  & 1,025,942 & 314,140  \\
3rd    & 670,319   & 354.423  & 1,036,973 & 207,247 \\
4th    & 698,008   & 827,72  & 459,200 & 27,689 \\
5th    & 702,887   & 12,010  & 143,546 & 4,879 \\
6th    & 703,705   & 2,711  & 51,348 & 818 \\
7th    & 703,901   & 416  & 12,887  & 196\\
8th    & 703,937   & 52  & 1,341 & 36 \\
9th    & 703,938   & 1  & 12  & 1\\
10th    & 703,938   & 0  & 0  & 0\\

\bottomrule
\end{tabular}
\caption{Spelling dictionary expansion iterations details, showing for each iteration the size of the spelling dictionary, the number of extracted OSCAR texts and distinct words, and the size difference increase of the spelling dictionary.}\label{iteration_details}%
\end{table*}

\subsection{GPT GEC}

The emergence of ChatGPT \citep{ouyang2022training} has significantly impacted the field of Natural Language Processing (NLP), marking the start of a new era of language generation and understanding. ChatGPT, based on OpenAI's GPT architecture, has demonstrated remarkable capabilities in generating human-like responses to text inputs across various domains. Its versatile applications span from conversational agents and chatbots to content generation, summarization, and translation tasks. 

Moreover, researchers and developers leverage ChatGPT as a benchmark for evaluating language understanding and generation models \citep{wang2023chatgpt}. It has also been used in many cases as an annotation means to annotate unlabeled data \citep{gilardi2023chatgpt, zhu2023can}, which we do in this work. We randomly sample 100k sentences from our OSCAR GEC dataset and prompt ChatGPT to correct the incorrect sentences, generating a parallel GPT GEC dataset of 100k parallel incorrect-correct sentences. 

\subsection{Datasets Overview}

Table \ref{tab2} presents a summary of the datasets employed in this study. We utilize three datasets, including two internally developed ones named OSCAR GEC and GPT GEC, which we introduced in previous sections, as well as an open-source dataset called GECTurk \citep{kara2023gecturk}, which we compare with our datasets.

\begin{table*}[h]
\centering
\begin{tabular}{|c|c|c|c|}
\toprule
\textbf{Error Type} & \textbf{OSCAR GEC} & \textbf{Turkish Tweets} & \textbf{Movie Reviews} \\
\toprule
SPELL & 0.4442 & 0.5175 & 0.0925 \\
ORTH & 0.1131 & 0.2579 & 0.5727 \\
OTHER & 0.1441 & 0.1116 & 0.1894 \\
NOUN & 0.0160 & 0.0155 & 0.0529 \\
NOUN:INFL & 0.0180 & 0.0108 & 0.0044 \\
NOUN:NUM & - & 0.0026 & 0.0044 \\
PRON & 0.0014 & 0.0037 & - \\
VERB:INFL & 0.0133 & 0.0363 & 0.0088 \\
ADJ & 0.0059 & 0.0077 & 0.0220 \\
    CONJ & 0.0048 & 0.0103 & 0.0352 \\
NUM & 0.0047 & - & - \\
DET & 0.0020 & 0.0019 & - \\
QUES & 0.0008 & 0.0015 & 0.0044 \\
ADJ:POSS & 0.0003 & - & - \\
ADJ-VERB:INFL:POSS & 0.0004 & 0.0002 & - \\
ADJ-VERB:INFL:CASE & 0.0000 & - & - \\
ADV-VERB:INFL:CASE & 0.0001 & - & - \\
ADV & 0.0043 & 0.0065 & - \\
PUNC & 0.2070 & 0.0002 & 0.0132 \\
VERB:SVA & 0.0021 & 0.0011 & - \\
VERB & 0.0064 & 0.0071 & - \\
PREP & 0.0040 & 0.0044 & - \\
NOUN-VERB:INFL:POSS & 0.0001 & - & - \\
VERB:TENSE & 0.0022 & 0.0005 & - \\
WO & 0.0014 & - & - \\
\bottomrule
\end{tabular}
\caption{Error Types and their percentages in the evaluation sets mentioned in Table \ref{tab2} classified by ERRANT-TR}\label{eval_sets_error_types}%
\end{table*}

As for evaluation, we randomly pick 2,408 sentences from our OSCAR GEC dataset, which we exclude from training and validation, and manually annotate them forming a manually annotated evaluation set of organic 2,408 sentences. We also test and benchmark on open-source evaluation sets such as \citep{koksal-etal-2020-turki}, which contains 1,996 tweets, and \citep{kara2023gecturk}'s curated evaluation set which includes 300 movie reviews. In Table \ref{eval_sets_error_types}, we show the percentages of the frequencies of the error types found in the three evaluation sets OSCAR GEC (22,366 errors), Turkish Tweets (6,201 errors), and Movie Reviews (227 errors) classified by ERRANT-TR. The Table shows that the OSCAR GEC evaluation set has more error types than the other two.

\begin{table*}[h]
\centering
\begin{tabular}{p{4.8cm}p{1cm}p{1.3cm}p{1cm}p{1.8cm}p{1.95cm}}
\toprule
Dataset Name & Split & Sentences & Tokens & Error Types & Domain \\

\midrule
OSCAR GEC (ours) & Train &  2.3m & 213.2m & ERRANT &  Web \\
GPT GEC (ours) & Train &  100k & 3.6m & ERRANT &  Web \\
GECTurk \citep{kara2023gecturk} & Train &  138k & 5.8m& 25 & Newspapers \\
\midrule
OSCAR GEC (ours) & Test &  2.4k & 142k & ERRANT &  Web \\
Movie Reviews \citep{kara2023gecturk} & Test &  300 & 2.7k & 25 &  Movie Reviews \\
Turkish Tweets \citep{koksal-etal-2020-turki} & Test &  2k & 116.2k & 13 &  Tweets \\
\bottomrule
\end{tabular}
\caption{An overview of the datasets utilized in this work. The datasets in the top half are synthetic and the bottom ones, the evaluation sets, are humanly annotated. The error type ERRANT refers to the automatic annotation tool ERRANT \citep{bryant2017automatic, felice2016automatic}, which automatically annotates parallel sentences with error-type information. Tokens information is based on OpenAI's tokenizer tiktoken with gpt2 encodings}\label{tab2}%
\end{table*}

\section{Experimental Setup}\label{sec4}

\subsection{Models}

We perform several experiments in this work with the mT5 model \cite{xue2020mT5}, a multilingual pre-trained encoder-decoder text-to-text transformer trained on a Common Crawl-based dataset covering 101 languages. We fine-tune the model on three different datasets: the OSCAR GEC, GPT GEC, and GECTurk \cite{kara2023gecturk}. Even though we train an mT5 model on the GECTurk dataset, we compare our mT5 models to their model, which is a sequence tagger based on a pre-trained Turkish cased Bert model \cite{stefan_schweter_2020_3770924} with extra linear and softmax layers similar to the work done in \cite{omelianchuk-etal-2020-gector}.

We train our models on one NVIDIA GeForce RTX 3090 for 10 epochs while only saving the best three checkpoints. Since some of the sentences extracted from OSCAR are long, we truncate the sentences to a max length of 48.

\subsection{Evaluation}

We evaluate our models and GECturk's sequence tagger on three different evaluation sets listed in Table \ref{tab2}: OSCAR GEC, Movie Reviews, and Turkish Tweets. To automatically annotate the parallel sentences of our evaluation sets and model outputs, we use ERRANT-TR \cite{uz-eryigit-2023-towards}, a variant of ERRANT \cite{bryant2017automatic, felice2016automatic} developed for the Turkish language. Table \ref{table:errant_errors} shows the error types, descriptions, and examples defined in the original ERRANT framework, while Table \ref{tab:errantTR_errors} shows the mapped error types in the ERRANT-TR framework.

ERRANT-TR outputs the annotations in M2 format \cite{dahlmeier2012better}, which can be evaluated using ERRANT's evaluation scripts that calculate the F0.5 score given two M2 files: a reference file (i.e. the gold M2 file) and a hypothesis file (i.e. the model output M2 file). ERRANT provides different scoring modes such as span-based correction, span-based detection, and token-based detection. 

To sum things up, we follow the following steps in evaluating our models: 1) Generate a parallel tab-separated file of the source-gold sentences. 2) Generate a parallel tab-separated file of the source-model\_output sentences. 3) Generate M2 files for the previously mentioned two tab-separated files. 4) Calculate the score of each model on every evaluation set given the corresponding gold and model\_output M2 files. 

We post-process the Turkish Tweets evaluation set in Table \ref{tab2} and the model outputs of it for a fair evaluation and comparison. We apply two transformations to it. First, to the evaluation set: 1) We capitalize the first letter of each correct sentence in the dataset since most models are trained with the data that way. Second, to the model outputs: 2) We remove the punctuation marks from the model outputs since they were removed from the evaluation set \cite{koksal-etal-2020-turki}. We apply the previous two transformations for all models making sure our results and comparisons are accurate.

\section{Results}\label{sec5}

Table \ref{tab3} and Table \ref{tab4} show the precision, recall, and F0.5 scores of correction and detection, respectively, for all models tested on three different test sets. We evaluate four different models on every test set. GECTurk (Seq Tagger), is a sequence tagger trained by \cite{kara2023gecturk} on their GECTurk training set. We also fine-tune mT5 on the same dataset, which is the model GECTurk (mT5) in the table. The remaining two models GPT GEC and OSCAR GEC are also fine-tuned mT5 models on our GPT GEC and OSCAR GEC training sets respectively. 

\begin{table*}[h]
\centering
\begin{tabular}{p{6.5cm}p{1cm}p{1cm}p{1cm}}
\toprule
Model & P & R & F0.5 \\

\midrule
\multicolumn{4}{c}{\textbf{OSCAR GEC} (ours)} \\
\midrule
GECTurk (Seq Tagger) \cite{kara2023gecturk} & 49.0 & 3.9 & 14.7\\
GECTurk (mT5) & 42.5 & 5.7 & 18.2\\
\hline
GPT GEC (mT5) & \textbf{69.8} &  \textbf{44.9} & \textbf{62.8} \\
OSCAR GEC (mT5) & 68.7 & 31.2 & 55.4 \\

\midrule
\multicolumn{4}{c}{\textbf{Turkish Tweets} \cite{koksal-etal-2020-turki}} \\
\midrule

GECTurk (Seq Tagger) \cite{kara2023gecturk} & 64.7 &  19.8 & 44.5\\
GECTurk (mT5) & 57.2 &  20.7 & 42.3\\
\hline
GPT GEC (mT5) & 77.7 &  \textbf{68.9} & 75.8 \\
OSCAR GEC (mT5) & \textbf{85.1} &  61.3 & \textbf{79.0} \\

\midrule
\multicolumn{4}{c}{\textbf{Movie Reviews} \cite{kara2023gecturk}} \\
\midrule

GECTurk (Seq Tagger) \cite{kara2023gecturk} & \textbf{86.5} &  \textbf{76.2} & \textbf{84.2}\\
GECTurk (mT5) & 73.1 &  71.8 & 72.8\\
\hline
GPT GEC (mT5) & 36.0 &  46.3 & 37.6 \\
OSCAR GEC (mT5) & 30.0 & 22.5 &  28.1\\

\bottomrule

\end{tabular}
\caption{ERRANT span-based \textbf{correction} scores (precision, recall, and F0.5) of every model on all the evaluation sets. All mT5 are trained are ours. GECTurk (Seq Tagger) is trained by the referenced work. The highest value of all models on each metric is bolded per evaluation set.}\label{tab3}
\end{table*}

\subsection{Correction}

In Table \ref{tab3}, which shows the ERRANT span-based correction scores, we notice the poor performance, on the OSCAR GEC test set, of both models trained on the GECTurk dataset achieving an F.05 score of 14.7 by the Sequence Tagger and 18.2 by the fine-tuned mT5. Surprisingly, those models' recall is significantly lower than the recall of the other two models being at most 5.7, which can be interpreted by the fact that the GECTurk dataset covers only 25 error types of all possible error types. Besides, the fact that the fine-tuned mT5 on the GECTurk dataset is doing better than the sequence tagger supports our claim that relying on the knowledge in large models such as mT5 is beneficial. For example, the GECTurk (mT5) model transforms the word Yuzune (to your face), which contains a deasciification error, into Yüzüne, the correct version despite the fact this error type is missing in the GECTurk dataset. On the other hand, GPT GEC and OSCAR GEC models perform significantly better with an F0.5 score of 62.8 and 55.4 respectively. 

\begin{figure*}[h]
	\centering
	\includegraphics[width=.8\textwidth]{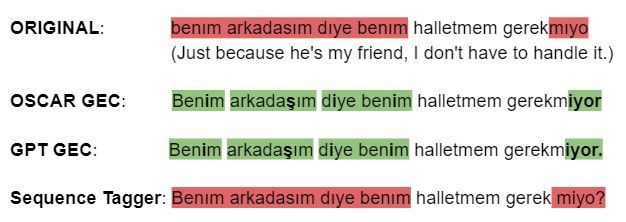}
	\caption{One example from the Turkish Tweets and the output of the three models OSCAR GEC, GPT GEC, and Sequence Tagger. The red segments are incorrect and the green ones are correct.}
	\label{example_outputs}
\end{figure*}

For the Turkish Tweets test set, the OSCAR GEC model is achieving the highest F0.5 score of 79.0, slightly higher than the GPT GEC model, and significantly higher than the models trained on the GECTurk dataset, which score at most an F0.5 score of 44.5. One of the reasons the OSCAR GEC model is slightly better than the GPT GEC is that GPT sometimes considered hashtags as grammatical mistakes in its annotations and replaced them. Perhaps prompting with this in mind could help in overcoming this problem. We show an example from the Turkish Tweets evaluation set in Figure \ref{example_outputs}. The OSCAR GEC and GPT GEC models correct the text with a slight difference in punctuation. The sequence Tagger from \cite{kara2023gecturk}, on the other hand, leaves half of the sentence incorrect and even corrupts the last word. 

On the Movie Reviews evaluation set, which is annotated by the same authors who built the GECTurk training set, the GECTurk models achieve noticeably a higher F0.5 score, 84.2 by the Sequence Tagger, than the GPT GEC and OSCAR GEC models which at most score an F0.5 score of 37.6. One of the reasons our models score low here is the annotation inconsistencies in the Movie Reviews evaluation set. For instance, the GPT GEC model capitalizes people's names such as Matt Damon while the evaluation set's gold annotations are sometimes capitalized and sometimes left in lowercase. Another reason is that the evaluation set has wrong annotations such as Matrixten (from the Matrix) which the GPT GEC model corrects as Matrix'ten, its correct version. Another wrong annotation example is the word orjinalinde (in the original one) that GPT GEC corrects as orijinalinde, which is its correct version. 

\subsection{Detection}

\begin{table*}[h]
\centering
\begin{tabular}{p{6.5cm}p{1cm}p{1cm}p{1cm}}
\toprule
Model & P & R & F0.5 \\

\midrule
\multicolumn{4}{c}{\textbf{OSCAR GEC} (ours)} \\
\midrule
GECTurk (Seq Tagger) \cite{kara2023gecturk} & 79.2 &  6.2 & 23.7\\
GECTurk (mT5) & 73.0 &  9.6 & 31.3\\
\hline
GPT GEC (mT5) & \textbf{85.7} &  \textbf{55.1} & \textbf{77.1} \\
OSCAR GEC (mT5) & 82.1 & 37.3 & 66.2 \\

\midrule
\multicolumn{4}{c}{\textbf{Turkish Tweets} \cite{koksal-etal-2020-turki}} \\
\midrule

GECTurk (Seq Tagger) \cite{kara2023gecturk} & 90.5 &  27.6 & 62.2\\
GECTurk (mT5) & 85.4 &  30.9 & 63.1\\
\hline
GPT GEC (mT5) & 92.0 &  \textbf{81.7} & \textbf{89.7} \\
OSCAR GEC (mT5) & \textbf{95.3} &  68.5 & 88.4 \\

\midrule
\multicolumn{4}{c}{\textbf{Movie Reviews} \cite{kara2023gecturk}} \\
\midrule

GECTurk (Seq Tagger) \cite{kara2023gecturk} & \textbf{90.5} &  \textbf{79.7} & \textbf{88.1}\\
GECTurk (mT5) & 78.5 &  77.1 & 78.2\\
\hline
GPT GEC (mT5) & 43.2 &  55.5 & 45.2 \\
OSCAR GEC (mT5) & 34.1 &  25.6 & 32.0 \\

\bottomrule

\end{tabular}
\caption{ERRANT span-based \textbf{detection} scores (precision, recall, and F0.5) of every model on all the evaluation sets. All mT5 are trained are ours. GECTurk (Seq Tagger) is trained by the referenced work. The highest value of all models on each metric is bolded per evaluation set}\label{tab4}%
\end{table*}

The detection results in Table \ref{tab4} mostly follow the trend of the correction results. On the OSCAR GEC test set, the GPT GEC model scores the highest F0.5 score of 77.1. Besides, the GECTurk models also perform poorly here with at most an F0.5 score of 31.3. The gap between the correction and detection F0.5 scores is high because, for example, the GPT GEC makes word-choice corrections that aren't wrong but unnecessary or missing in the gold annotations.  

On the Turkish Tweets evaluation set, GPT GEC and OSCAR GEC models also significantly do better than GECTurk models with scores F0.5 scores of 89.7 and 88.4 respectively. While the GECTurk models, the Sequence Tagger, and the fine-tuned mT5, only scored 62.2 and 63.1 respectively. The OSCAR GEC model here, again, is more precise than the GPT GEC model because, for example, it does not make as many corrections, e.g. contextual ones, as GPT GEC does.

Finally, the GECTurk models again do well on the Movie Reviews evaluation set with at most an F0.5 score of 88.1 and at least a score of 78.2 by the Sequence Tagger and the fine-tuned mT5 models respectively. GPT GEC and OSCAR GEC, on the other hand, struggle again in detection with at most an F0.5 score of 45.2 by the Sequence Tagger. Again, the OSCAR GEC and the GPT GEC models struggle here for many reasons such as the ones we mentioned in the previous section. 

\section{Language Models (LMs)}\label{sec6}

\begin{table*}[h]
\centering
\begin{tabular}{p{6cm}p{1.5cm}p{1.5cm}}
\toprule
Model & Train Loss & Val Loss \\

\midrule
\multicolumn{3}{c}{\textbf{Original sentences} + Turkish OSCAR sample} \\
\midrule

GPT-2 (30M) & 2.38 & 2.39 \\
GPT-2 (124M) & 1.85 & 2.11 \\

\midrule
\multicolumn{3}{c}{\textbf{Corrected sentences} + Turkish OSCAR sample} \\
\midrule

GPT-2 (30M) & \textbf{2.26} & \textbf{2.28} \\
GPT-2 (124M) & \textbf{1.77} & \textbf{2.01} \\

\bottomrule

\end{tabular}
\caption{Trainig and validation losses of 4 different GPT-2 models of two different sizes trained on two dataset combinations having a common Turkish OSCAR sample and either the original sentences of our OSCAR GEC dataset or the corrected sentences of our OSCAR GEC dataset cleaned using our clean insertions method. Bolded values show lower training and validation losses.}\label{gpt_training}%
\end{table*}

The rise of large language models marks a transformative era in artificial intelligence and natural language processing. These models, such as GPT-3 \cite{brown2020language} and LLaMA\cite{touvron2023llama} have been given significant attention due to their impressive capabilities in generating human-like text and performing various language-related tasks.

Large language models require billions of tokens, one of the bottlenecks of training large language models for low-resource languages, to achieve high performance on NLP tasks. Most large language models depend on web-based multilingual resources and datasets such as CommonCrawl, C4 \cite{raffel2020exploring}, and OSCAR. 

Unfortunately, web-crawled datasets are noisy and therefore need to be cleaned. For instance, we find that Turkish OSCAR\footnote{https://huggingface.co/datasets/oscar/viewer/unshuffled\_deduplicated\_tr}, which has around 11.6 million text documents, has at least one spelling mistake in 10\% of the documents based on our spelling dictionary.

In this chapter, we train four language models with different settings to show the effectiveness of our clean-insertions method for language models.

\begin{table*}[h]
\centering
\begin{tabular}{p{2.5cm}p{1cm}p{1cm}p{1cm}p{1cm}p{1cm}}
\toprule
Model & A1 & A2 & A3 & A4 & A5 \\

\midrule
\multicolumn{6}{c}{\textbf{Original sentences} + Turkish OSCAR sample} \\
\midrule

GPT-2 (30M) & \textbf{3.6} & \textbf{2.84} & \textbf{3.76} & \textbf{3.12} & \textbf{3.74} \\
GPT-2 (124M) & 2.96 & \textbf{2.94} & \textbf{3.82} & \textbf{3.36} & 3.54\\

\midrule
\multicolumn{6}{c}{\textbf{Corrected sentences} + Turkish OSCAR sample} \\
\midrule

GPT-2 (30M) & 3.06 & 2.78 & 3.58 & 2.64 & 3.44\\
GPT-2 (124M) & \textbf{3.1} & 2.74 & 3.68 & 2.96 & \textbf{3.74}\\

\bottomrule

\end{tabular}
\caption{The average ratings of 50 generated texts sampled per model. The samples are rated from 1 to 5  by five annotators (A1-A5).}\label{gpt_eval}%
\end{table*}

\subsection{Data Processing}

We use our OSCAR GEC parallel dataset which contains around 113.9 million training tokens in the original sentences, and 117.2 million training tokens in the parallel corrected sentences. We add a sample of around 2 million text documents from the untouched Turkish OSCAR samples to the original sentences and the parallel corrected sentences. With the added sample, we end up with two different datasets: 1) the original sentences and the OSCAR sample (305 million training tokens) and 2) the corrected sentences and the OSCAR sample (314.4 million training tokens). All token information is obtained from Karpathy's GPT-2 implementation code\footnote{https://github.com/karpathy/nanoGPT}.

\subsection{Training}

We train four different GPT-2 models, following Karpathy's GPT-2 implementation, with two different sizes: 30M and 124M. We train the models on one NVIDIA GeForce RTX 3090. We train the 30M size model for 300k iterations, and the 124M size model for 8k iterations. Table \ref{gpt_training} shows the training and validation losses. We notice lower training and validation losses for both sizes for the models trained on the corrected sentences and the Turkish OSCAR sample. This indicates the effectiveness of cleaning the Turkish OSCAR dataset with clean insertions using our spelling dictionary on the training losses. Maybe to be certain of this effect, we need to train different architectures other than GPT-2, which we leave as a future work.

\subsection{Evaluation}

We also evaluate our GPT models manually to see if there is an effect on the generated text. We generate 50 samples for every model using 50 common Turkish-word prompts and ask 5 evaluators to evaluate the total 200 samples from 1 to 5 based only on their cohesiveness, ignoring the spelling mistakes in the generated text since we observe that the models trained with the misspelled words are already biased towards generating misspelled text.

Table \ref{gpt_eval} shows the average rating by each annotator, from A1 to A5 for every model. As we see in the table, the clean insertions method did not lead to higher ratings for all models despite its obvious effect on the training losses. This could be because this is a small experiment with only 50 prompts. Or, maybe corrupting misspellings, i.e. correcting them, this way corrupts the context or causes imbalances in the training data.

\section{Conclusion}\label{sec13}

In summary, this study addresses the lack of attention within the research community to the Turkish Grammatical Error Correction task by introducing a method, clean insertions, that helps in creating organic Turkish GEC datasets. We open-source several datasets two of which are training datasets and one evaluation set. In addition to that, we share our models that achieve state-of-the-art results on two evaluation sets out of the three available evaluation sets. 

Our method, clean insertions, is simple to understand and apply. Other than the starting spelling dictionary that we build manually, it is fully automated. Normally, a synthetic dataset requires clean data to start with, which may not be available, however, our method works with any organic data, which usually contains grammatical errors. This leads to datasets that contain various types of errors and not only a set of pre-defined injected error types, which could cause the models trained on such datasets to perform poorly on evaluation sets containing error types out of the pre-defined set as we show in section \ref{sec5}. 

While our method yields partially correct parallel GEC datasets, since the spelling dictionary would not contain all possible errors, it can be used to obtain state-of-the-art results by relying on the knowledge in the large pre-trained models such as mT5. This finding is surprising and raises the question of whether we can solve other tasks the same way with partially correct or partially correctly annotated datasets. Certainly, such datasets would confuse the models in tasks such as text classification, but it is maybe worth trying for tasks that can be formulated as text-to-text problems. 

In addition to the dataset we build using clean insertions, OSCAR GEC, we use GPT as an annotator and build a GEC dataset, GPT GEC, to show the potential of using such models as annotators. Indeed, the models trained on the GPT GEC show promising results surpassing the other models on most evaluation sets as we show in Table \ref{tab3} and \ref{tab4}.

While this study provides valuable insights, it is important to acknowledge its limitations, such as the need for a manually annotated spelling dictionary to start with. Another limitation is the lack of a clear and specific error-type annotation. Since we use ERRANT, which automatically classifies the errors, the set of error types is limited and not detailed and specialized enough for the Turkish language.  

Future work could focus more on using other or even more complex and context-aware components in addition to the Spelling Checker and Deasciifier we utilize in our OSCAR GEC pipeline. With more components, more incorrect-correct pairs would be added to the spelling dictionary, which could lead to higher-quality datasets. Besides, applying the approach to datasets other than Turkish OSCAR could enrich the OSCAR GEC dataset with examples that contain missing error types in the current dataset.

\bibliography{anthology,custom}
\bibliographystyle{acl_natbib}
\newpage
\appendix

\section{ERRANT Error Types}
\label{sec:appendix1}

In this section, we show the error types pre-defined in the ERRANT framework and their mapped Turkish version. Table \ref{table:errant_errors} shows the error types, descriptions, and examples defined in the original ERRANT framework, while Table \ref{tab:errantTR_errors} shows the mapped error types in the ERRANT-TR framework

\begin{table*}[h]
\centering
\begin{tabular}{p{0.2\textwidth}p{0.3\textwidth}p{0.5\textwidth}}
\toprule
\textbf{Code} & \textbf{Meaning} & \textbf{Description / Example} \\
\midrule
ADJ & Adjective & big → wide \\
ADJ:FORM & Adjective Form & Comparative or superlative adjective errors. goodest →  best, bigger → biggest, more easy → easier  \\
ADV & Adverb & speedily → quickly\\
CONJ & Conjunction & and → but \\
CONTR & Contraction & n't → not \\
DET & Determiner & the → a \\
MORPH & Morphology & Tokens have the same lemma but nothing else in common. quick (adj)→ quickly (adv) \\
NOUN & Noun & person → people \\
NOUN:INFL & Noun Inflection & Count-mass noun errors. informations → information \\
NOUN:NUM & Noun Number & cat → cats \\
NOUN:POSS & Noun Possessive & friends → friend's \\
ORTH & Orthography & Case and/or whitespace errors. Bestfriend → best friend \\
OTHER & Other & Errors that do not fall into any other category (e.g. paraphrasing). at his best →  well, job → professional \\
PART & Particle & (look) in → (look) at \\
PREP & Preposition & of → at \\
PRON & Pronoun & ours → ourselves \\
PUNCT & Punctuation & !→. \\
SPELL & Spelling & genectic → genetic, color → colour \\
UNK & Unknown & The annotator detected an error but was unable to correct it. \\
VERB & Verb & ambulate → walk \\
VERB:FORM & Verb Form & Infinitives (with or without "to"), gerunds (-ing) and participles. to eat → eating, dancing → danced \\
VERB:INFL & Verb Inflection & Misapplication of tense morphology. getted → got, fliped → flipped \\
VERB:SVA & Subject-Verb Agreement & (He) have → (He) has \\
VERB:TENSE & Verb Tense & Includes inflectional and periphrastic tense, modal verbs and passivization. eats → ate, eats → has eaten, eats → can eat, eats → was eaten \\
WO & Word Order & only can → can only \\
\bottomrule
\end{tabular}
\caption{The list of 25 main error categories in the ERRANT framework with examples and explanations as listed in their work.}
\label{table:errant_errors}
\end{table*}

\begin{table*}[h]
\begin{tabular}{|p{2.5cm}|p{8cm}|p{4cm}|}
\hline
\textbf{Error Code} & \textbf{Meaning} & \textbf{Example} \\ \hline
ADJ & Wrong choice of adjective & büyük → küçük  \\ \hline
ADJ:FORM & Wrong usage of comparative or superlative adjective &  \\ \hline
ADV & Wrong choice of adverb & önce → sonra  \\ \hline
CONJ & Wrong choice of conjunction & ama → belki  \\ \hline
CONTR & Wrong choice of contraction &  \\ \hline
DET & Wrong choice of determiner & bu elma → o elma  \\ \hline
MORPH & Tokens have the same lemma but nothing else in common & kalem → silgi  \\ \hline
NOUN & Wrong choice of nouns &  \\ \hline
NOUN:INFL & Count-mass noun errors &  \\ \hline
NOUN:NUM & Wrong usage of noun number & elma → elmalar  \\ \hline
NOUN:POSS & Wrong usage of noun possessive & hastalarının ilaçları → hastaların ilaçları  \\ \hline
ORTH & Case and/or whitespace errors & herşey → her şey  \\ \hline
OTHER & Errors that do not fall into any other category &  \\ \hline
PART & Wrong choice of particle &  \\ \hline
PREP & Wrong choice of preposition & gibi → için  \\ \hline
PRON & Wrong usage of pronoun & sen → ben  \\ \hline
PUNCT & Wrong usage of punctuation & ? → !  \\ \hline
SPELL & Misspelling & broblem → problem  \\ \hline
UNK & A detected but not corrected error &  \\ \hline
VERB & Wrong choice of verbs & geldim → gittim  \\ \hline
VERB:FORM & Infinitives, gerunds and participles & gitmek, gitme, giden  \\ \hline
VERB:INFL & Wrong usage of tense morphology & (biz) yaptık → (biz) yaptık  \\ \hline
VERB:SVA & Subject-verb agreement & sen geliyorum → sen geliyorsun  \\ \hline
VERB:TENSE & Wrong choice of inflectional and periphrastic tense, modal verbs and passivization & geliyorum → gelmiştim  \\ \hline
WO & Word order & elma kırmızı → kırmızı elma  \\ \hline
\end{tabular}
\caption{ERRANT-TR's Error Codes, Descriptions, and Examples as they list in their work. An empty cell indicates that the category has no example of being either too wide or not useful for Turkish.}
\label{tab:errantTR_errors}
\end{table*}

\end{document}